# Identifying physical health comorbidities in a cohort of individuals with severe mental illness: An application of SemEHR

Rebecca Bendayan[1, 2], Honghan Wu,[3,4], Zeljko Kraljevic[1], Robert Stewart[2,5], Tom Searle[1], Jaya Chaturvedi[1], Jayati Das-Munshi[2,5], Zina Ibrahim[1], Aurelie Mascio[1], Angus Roberts[1,2], Daniel Bean[1,6], Richard Dobson[1,2,6,7]

[1]Department of Biostatistics and Health Informatics, Institute of Psychiatry, Psychology and Neuroscience, King's College London, London, United Kingdom
[2]NIHR Biomedical Research Centre at South London and Maudsley NHS Foundation Trust and King's College London, London, United Kingdom
[3]Centre for Medical Informatics, Usher Institute, University of Edinburgh, United Kingdom
[4]Health Data Research UK Edinburgh, University of Edinburgh, United Kingdom
[5]Department of Psychological Medicine, Institute of Psychiatry, Psychology and Neuroscience, King's College London, London, United Kingdom
[6]Health Data Research UK London, University College London, London, United Kingdom
[7]Institute of Health Informatics, University College London, London, United Kingdom

**Abstract**

*Multimorbidity research in mental health services requires data from physical health conditions which is traditionally limited in mental health care electronic health records. In this study, we aimed to extract data from physical health conditions from clinical notes using SemEHR. Data was extracted from Clinical Record Interactive Search (CRIS) system at South London and Maudsley Biomedical Research Centre (SLaM BRC) and the cohort consisted of all individuals who had received a primary or secondary diagnosis of severe mental illness between 2007 and 2018. Three pairs of annotators annotated 2403 documents with an average Cohen's κ of 0.757. Results show that the NLP performance varies across different diseases areas (F1 0.601 – 0.954) suggesting that the language patterns or terminologies of different condition groups entail different technical challenges to the same NLP task.*

**Introduction**

The Academy of Medical Sciences (2018) has highlighted that the increase of multimorbidity (2 or more co-existent health conditions) in our population constitutes a challenge for our health care. Research on multimorbidity traditionally focuses on the ageing population, however there is a need to understand multimorbidity in other populations such as individuals with mental health disorders. Similarly, the Framework for Mental Health Research developed by the Department of Health (2017) acknowledged the need to account for the interactions between mental and physical health to reduce the mortality gap between individuals with severe mental illnesses (SMI), such as schizophrenia or bipolar disorder, and the general population.

Large national population studies have very limited data for individuals with mental health disorders and therefore studies have to use other data sources such as electronic health records (EHRs). The rise of the use of EHRs and the UK government's commitment for the NHS to be paperless by 2020 provides us a unique opportunity to access relevant data for this population. One of the largest providers of secondary mental health care in UK and Europe is the South London and Maudsley (SLaM) NHS Foundation Trust. In 2007, the SLaM NIHR Biomedical Research Centre (BRC) developed the Clinical Record Interactive Search (CRIS) system to enable routinely collected mental health EHRs to be used in research, and since 2013 this has been deployed successfully at other mental health NHS Foundation Trusts across the UK. However, structured data on physical health conditions is limited in mental health care EHRs and this data is mainly hidden in unstructured clinical notes which has so far limited their use for the study of multimorbidity in SMI. Within this context, there is a need to make this data on physical health conditions available for researchers.

SemEHR, an open source toolkit that integrates text mining and semantic computing for identifying mentions of Unified Medical Language System (UMLS) concepts from clinical documents, has been particularly helpful in extracting data from systems such as the CRIS system [1]. A preliminary study [2] focusing in schizophrenia patients has used SemEHR to extract and validate 9 ICD-10 chapter level codes representing viral infection, endocrine, neurologic, cardiovascular, respiratory, digestive, skin, musculoskeletal and urogenital systems. The present study builds on this work as we aim to identify a larger number of chapter level codes and validate them in the larger SMI cohort, including additionally bipolar affective disorders and non-organic psychoses.

**Methods**

*Corpus selection and preprocessing: The South London and Maudsley Mental Health Case Register*

Data was extracted from CRIS at SLaM, a mental health provider to 1.2 million residents from south London which has one of the highest recorded incidence rates of psychosis in the UK and includes some of the largest ethnic minority communities in the UK such as Black Caribbean, Black African and South Asian communities [3]. Documents were extracted from all individuals aged 15 years or older who had received a primary or secondary diagnosis of severe mental illness (SMI) between 2007 and 2018 (according to the International Classification of Mental and Behavioural Disorders-10 (ICD-10). More specifically, this included schizophrenia-spectrum disorders, bipolar affective disorders and non-organic psychoses.

*Definition of physical health conditions and Information extraction*

Physical health conditions were ascertained from data reported in text records from CRIS during the 3 months before first SMI diagnosis was registered until 2018 for each individual. For this data extraction we used SemEHR [1] which has shown good recall (96-98%) for physical health conditions at document level in studies using SLaM [1]. Moreover, SemEHR is especially beneficial as it uses UMLS concepts instead of ICD-10 terms which allows us to identify as many variants of disease mentions as possible and contributes to disambiguation. To identify mentions of a broad range of physical diseases (A00-N99), we mapped each top level ICD code (3-character code, e.g. A00) to a corresponding UMLS concept CUI (e.g. C0008354) using the mappings at BioPortal by querying its SPARQL Endpoint.

*Information Annotation*

Generally, we defined a relevant instance as a mention of a physical health condition experienced by the patient that is recent and does not appear negated. In total, 2403 documents were annotated by 6 annotators using the MedCATTrainer interface [4] where the annotators had to specify whether the condition was correctly identified, negated or not and whether the mention was historic or recent. All annotators were previously trained for this task and guidelines are available upon request. These annotators were three clinicians (RB, RS, JC) and three non-clinicians (ZK, TS, AM), and inter-annotator agreement statistics were derived for a clinician-clinician pair and 3 non-clinician-clinician pairs. Out of the 2403 documents we have chosen 2084 where the annotators agree and created a gold standard dataset.

*Improving SemEHR results using labelled data*

We used SemEHR as the NLP tool to extract mentions of physical conditions from free text notes. As we showed in our previous work [5], SemEHR performance can be further improved when labelled data is available. The labelled data was sampled from SemEHR processed whole cohort and subsequently verified by human/domain experts. We therefore implemented an extra step of using a machine learning model to classify SemEHR identified entities as true condition mentions or not. A separate classifier model is trained which takes as input of all UMLS concepts identified by SemEHR within the sentence where the entity appears and the prior sentence. We chose a Random Forest classifier as it is one of the best performing models in a previous study on extracting phenotypes from radiology reports [5].

**Results**

*Inter-annotator agreement and model validation*

In total, 220 physical health conditions (distinct UMLS concepts) were chosen, and a total of 2084 instances of health conditions were annotated from 2084 documents to create gold standards and training data specific to each physical health conditions. These health conditions were double annotated, yielding an average Cohen's κ of 0.757 (Chapter Level presented in Table 1).

**Table 1.** Cohen´s κ for each ICD 10 Chapter Level.

| ICD10 chapter | I | II | III | IV | V | VI |
|---|---|---|---|---|---|---|
| Cohen's κ | 0.77 | 0.79 | 0.98 | 0.79 | 0.55 | 0.57 |
| **ICD10 chapter** | VII | IX | X | XI | XII | XIII |
| Cohen's κ | 0.55 | 0.76 | 0.80 | 0.74 | 0.70 | 0.56 |
| **ICD10 chapter** | XIV | XV | XVII | XVIII | XIX | XX |
| Cohen's κ | 0.50 | 0.75 | 0.87 | 0.78 | 0.68 | 0.65 |

On the above described annotated dataset, we did a 10 fold cross validation using a random forest classifier. Summary statistics (macro-average F1 scores) aggregated across all for ICD 10 Chapter Level are presented in Table 2.

**Table 2.** Macro-average Precision, Recall and F1-Score at ICD 10 Chapter Level using 10 Fold cross validation on the annotated dataset

| ICD10 chapter | I | II | III | IV | V | VI |
|---|---|---|---|---|---|---|
| Instances | 64 | 72 | 42 | 51 | 113 | 64 |
| Precision | 0.662 | 0.748 | 0.737 | 0.866 | 0.920 | 0.767 |
| Recall | 0.953 | 0.958 | 0.952 | 0.961 | 0.991 | 0.984 |
| F1 | 0.781 | 0.840 | 0.831 | 0.911 | 0.954 | 0.862 |
| **ICD10 chapter** | VII | IX | X | XI | XII | XIII |
| Instances | 83 | 107 | 99 | 138 | 138 | 80 |
| Precision | 0.802 | 0.858 | 0.879 | 0.907 | 0.873 | 0.904 |
| Recall | 0.952 | 0.916 | 1.000 | 0.964 | 0.993 | 0.975 |
| F1 | 0.870 | 0.886 | 0.936 | 0.935 | 0.929 | 0.938 |
| **ICD10 chapter** | XIV | XV | XVII | XVIII | XIX | XX |
| Instances | 65 | 82 | 31 | 124 | 58 | 46 |
| Precision | 0.885 | 0.874 | 0.479 | 0.816 | 0.682 | 0.610 |
| Recall | 0.969 | 0.976 | 0.806 | 0.952 | 0.897 | 0.826 |
| F1 | 0.925 | 0.922 | 0.601 | 0.878 | 0.775 | 0.702 |

**Discussion**

The performance varies across different chapters indicating that the language patterns or terminologies of different condition groups present different technical challenges to the same NLP task. For example, chapters X and XIV have >.90 F1 scores, while chapter XVII is only .60. Therefore, using a unified model to identify such a large number of conditions is probably not an ideal solution and better results may be obtained by training a model per chapter (or even per condition as for the classifier step). However, this approach would require substantially more training data.

In general, the NLP performance was not as optimal as in our previous study on radiology reports [5], where 0.91-0.95 F1 scores were achieved across 23 phenotypes on 3 different datasets. The first reason could be that it is a relatively easy task to identify phenotypes from mono-typed dataset (i.e. radiology reports only), where the input text is likely to be less varied compared to full mental health records. The second possible reason is that the labelled data is much smaller at condition level – only 7-8 labelled annotations per condition on average. Therefore, it was difficult to obtain a good classifier in many cases. Finally, it needs to be noted that reported performance in this paper is at the individual mention level and not at the patient level, i.e. our model does not determine whether a patient had a physical condition or not.

**Conclusion**

These results allow us to access information on physical health conditions for individuals with SMI and perform further patient-level analysis of multimorbidity. Specifically, for Chapters IV (Endocrine, Nutritional and Metabolic Diseases), V (Mental, Behavioral and Neurodevelopmental Disorders), X (Diseases of the respiratory system), XI (Diseases of the digestive system), XII (Diseases of the skin and subcutaneous tissue), XIII (Diseases of the musculoskeletal system and connective tissue) or XIV (Diseases of the genitourinary system). Overall, we provide a valuable resource for the growing body of research in multimorbidity using EHRs. As SemEHR has shown good transferability, future steps also involve 1) increasing more labelled data to achieve better models and 2) validating this process in other mental health NHS Foundation Trusts across the UK.